\documentclass[times,twocolumn,final]{elsarticle}

\usepackage{medima}
\usepackage{framed,multirow}

\usepackage{amssymb}
\usepackage{latexsym}

\usepackage{url}
\usepackage{xcolor}

\usepackage{hyperref}
\usepackage{amsmath,amssymb,amsfonts,booktabs,diagbox}
\usepackage[labelfont=bf,textfont=normalfont]{subcaption}
\usepackage[labelfont=bf,textfont={bf,bf}]{caption}
\usepackage{flushend}
\journal{Medical Image Analysis}

\begin{document}

\verso{T. Cai \textit{et~al.}}

\begin{frontmatter}

\title{\textit{DeepStroke}: An Efficient Stroke Screening Framework for Emergency Rooms with Multimodal Adversarial Deep Learning}

\author[1]{Tongan \snm{Cai}\corref{cor1}}
\ead{cta@psu.edu}
\author[1]{Haomiao \snm{Ni}\corref{cor1}}
\cortext[cor1]{T. Cai and H. Ni made equal contributions. All correspondence should be addressed to T. Cai, J.Z. Wang, and S.T.C. Wong.}
\author[2]{Mingli \snm{Yu}}
\author[1]{Xiaolei \snm{Huang}}
\author[3]{Kelvin \snm{Wong}}
\author[4]{John \snm{Volpi}}
\author[1]{James Z. \snm{Wang}}
\ead{jwang@psu.edu}
\author[3]{Stephen T.C. \snm{Wong}}
\ead{stwong@houstonmethodist.org}


\address[1]{College of Information Sciences and Technology, The Pennsylvania State University, University Park, Pennsylvania 16803, USA}
\address[2]{Department of Computer Science and Engineering, The Pennsylvania State University, University Park, Pennsylvania 16803, USA}
\address[3]{T.T. and W.F. Chao Center for BRAIN \& Houston Methodist Cancer Center, Houston Methodist Hospital, Houston, Texas 77030, USA}
\address[4]{Eddy Scurlock Comprehensive Stroke Center, Department of Neurology, Houston Methodist Hospital, Houston, Texas 77030, USA}

\received{8 Sept 2021}
\communicated{Albert C. S. Chung}

\begin{abstract}
In an emergency room (ER) setting, stroke triage or screening is a common challenge. A quick CT is usually done instead of MRI due to MRI's slow throughput and high cost. Clinical tests are commonly referred to during the process, but the misdiagnosis rate remains high. We propose a novel multimodal deep learning framework, \textit{DeepStroke}, to achieve computer-aided stroke presence assessment by recognizing patterns of minor facial muscles incoordination 
and speech inability for patients with suspicion of stroke in an acute setting. 
Our proposed \textit{DeepStroke} takes one-minute facial video data and audio data readily available during stroke triage for local facial paralysis detection and global speech disorder analysis. 
{Transfer learning was adopted to reduce face-attribute biases and improve generalizability.} We leverage a multi-modal lateral fusion to combine the low- and high-level features and provide mutual regularization for joint training. Novel adversarial training is introduced to obtain identity-free and stroke-discriminative features.
Experiments on our video-audio dataset with actual ER patients
show that \textit{DeepStroke} outperforms state-of-the-art models and achieves better performance than both a triage team and ER doctors, attaining a 10.94\% higher sensitivity and maintaining 7.37\% higher accuracy than traditional stroke triage when specificity is aligned. 
Meanwhile, each assessment can be completed in less than six minutes, demonstrating the framework's great potential for clinical translation. 
\end{abstract}


\end{frontmatter}

\newpage
\section{Introduction}
Stroke is a common cerebrovascular disease that can cause lasting brain damage, long-term disability, or even death~\citep{about}. It is the second leading cause of death and the third leading cause of disability worldwide~\citep{johnson2016stroke}. According to the Centers for Disease Control and Prevention~(\citeyear{cdc}), someone in the United States has a stroke every forty seconds and someone dies of a stroke every four minutes. 
In acute ischemic stroke where brain tissue lacks blood supply, the shortage of oxygen needed for cellular metabolism quickly causes long-lasting tissue damage. If identified and treated in time, many interventions are available and an acute ischemic stroke patient will have a greater chance of survival and subsequently a better quality of life. 

However, treatment delays related to misdiagnosis and underdiagnosis are common during triage~\citep{delay}. 
{A major problem in acute ischemic stroke is misdiagnosis. Misdiagnosis in stroke leads to undertreatment, overtreatment, and the potential for mental and physical disability~\citep{crichton2016patient}. Approximately 22\% of all ischemic stroke patients are missed during the pre-hospital triage screening, and the problem is more severe in community hospitals than in academic hospitals~\citep{arch2016missed}.}

Rapid diagnosis of acute ischemic stroke relies on clinical diagnosis and imaging as there is no point-of-care test available. Currently, the gold standard test for stroke is advanced neuro-imaging including diffusion-weighted MRI scan (DWI) that detects brain infarct with high sensitivity and specificity. Although accurate, DWI is usually not accessible in the emergency room (ER) due to limited equipment availability and high operating cost. Even in advanced centers with ER availability of MRI, the turnaround time for patient transport, testing, and results adds 30-60 minutes{, which is too inefficient for the stroke triage screening process. The healthcare system, instead, relies on rapid imaging such as CT and clinical judgment of nurses and physicians to detect neurological symptoms during emergency triage.} In the actual ER scenario, clinicians commonly adopt the following three tests:
the Cincinnati Pre-hospital Stroke Scale (CPSS)~\citep{kothari1999cincinnati}, the Face Arm Speech Test (FAST)~\citep{harbison2003diagnostic}, and the National Institutes of Health Stroke Scale (NIHSS)~\citep{nihss}. All these methods assess the presence of any unilateral facial droop, arm drift, and speech disorder. The patient is requested to repeat a specific sentence (CPSS) or have a conversation with the doctor (FAST), and abnormalities arise when the patient slurs, fails to organize his speech, or is unable to speak. For NIHSS, face and limb palsy conditions are also evaluated. However, the scarcity of neurologists~\citep{leira2013growing} makes such tests difficult to be timely and effectively conducted in all stroke emergencies. The evaluation may also fail to detect stroke cases where only very subtle facial motion deficits exist---that clinicians are unable to observe.

Recently, with the help of machine intelligence, researchers have proposed more and more accurate detection and evaluation methods for neurological disorders. Besides working on computer-aided medical image understanding for CT and MRI images~\citep{xue2018segan,YAO201814,akkus2017deep}, many researchers are now focusing on alternative contactless, efficient, and economic ways for the analysis of various neurological conditions. One of the most popular domains is the detection of facial paralysis with computer vision, {\it i.e.}, letting machines detect the anomalies in the subject's face. {However, the clinical scenario is always overlooked where obvious stroke patients are readily identified without treatment delay and subtle/non-obvious strokes are easily missed.  Obvious strokes can be identified by key-point methods~\citep{parra2021facial,zhuang2021video} but the same strategy will not be effective on subtle strokes.}

Moreover, the majority of work neglects the readily available and indicative speech audio features~\citep{facemedsurvey}, which can be an important source of information in stroke diagnosis. Also, current methods ignore the spatiotemporal continuity of facial motions~\citep{SLLE,morphing,anthropometry2015} and fail to tackle the problem of static/natural asymmetry. Common video classification frameworks like I3D~\citep{i3d} and SlowFast~\citep{slowfast} also fail to serve the stroke pattern recognition purpose due to the lack of training data and quick overfitting as ``subject-remembering'' effect. 

Worse still, few datasets of high quality have been constructed in the stroke diagnosis domain. The current clinical datasets \citep{6dsz-7d76-19,Kihara2011,CVPR2018,MEEI} are small (with hundreds of images or dozens of videos) and unable to comprehensively represent the diversity in stroke patients in terms of gender, race/ethnicity, and age. Also, they either evaluate normal subjects versus those with clear signs of a stroke~\citep{Bandini2016,gabor2016} or deal with full synthetic data (\textit{i.e.}, healthy people pretend to have palsy facial patterns)~\citep{isbi2018,umcnn2018}; some others establish experimental settings with hard constraints on the patient's head~\citep{Bevilacqua2011,Hamm2011,wang2014}. All these shortcomings will hinder their clinical implementation for ER screening or patient self-assessment. 

In this paper, we propose a novel deep learning framework, named \textit{DeepStroke}, {to assist the stroke triage team} in accurately and efficiently analyzing the presence of stroke in patients {admitted to the Emergency Rooms (ER) with a subtle sign of stroke.} The problem is formulated as a binary classification task, {\it i.e.}, stroke \textit{vs.} non-stroke. Instead of taking a single-modality input, the core network in \textit{DeepStroke} consists of two temporal-aware branches, the video branch for local facial motion analysis and the audio branch for global vocal speech analysis, to collaboratively detect the presence of stroke patterns. A novel lateral fusion scheme between these two branches is introduced to combine the low- and high-level features and provide mutual regularization for joint training. To mitigate the ``subject-remembering'' effect, \textit{DeepStroke} also adopts adversarial learning to extract identity-free and stroke-discriminative features, as well as transfer learning to reduce facial-attribute biases and improve the network generalizability.  

\begin{figure*}[htbp]
    \centering
    \includegraphics[page=1,width=.77\linewidth]{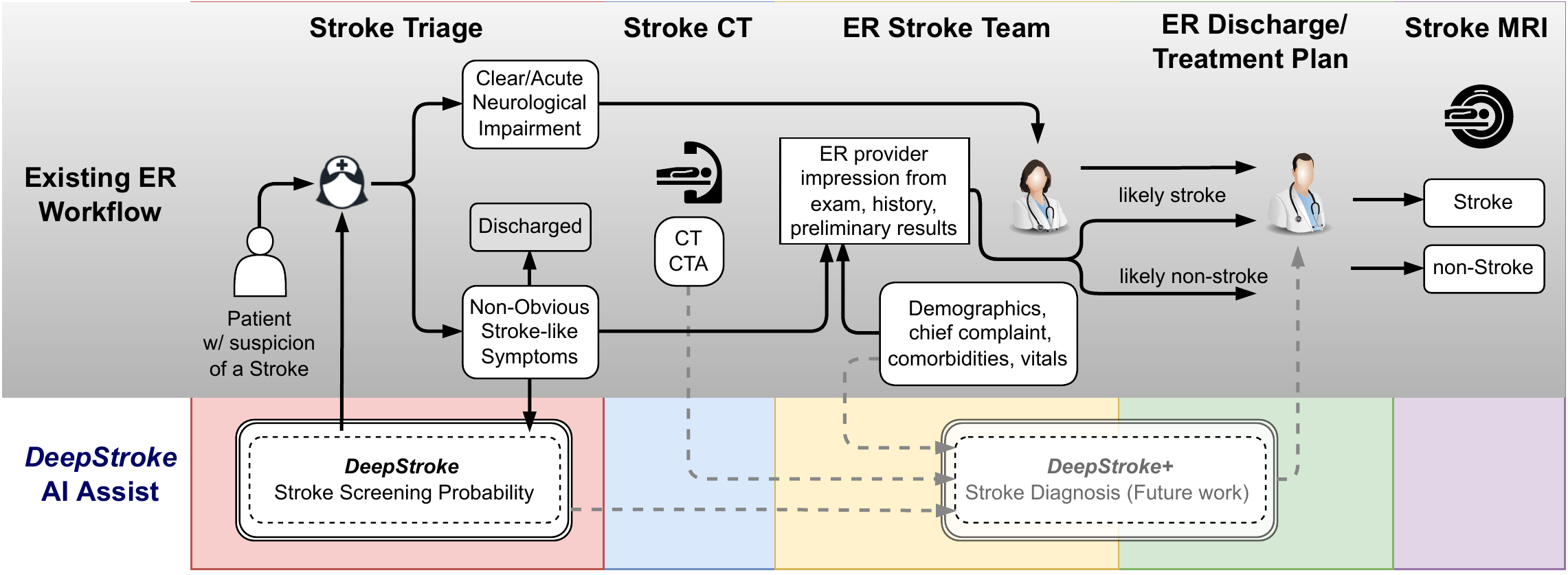}
    \caption{{The anticipated workflow of \textit{DeepStroke} in assisting stroke diagnosis in the ERs. }}
    \label{fig:wf}
\end{figure*}

{To better illustrate the use case of our proposed method, we plot the anticipated clinical workflow of \textit{DeepStroke} in helping stroke diagnosis in ER, as shown in Fig.~\ref{fig:wf}. If the incoming patient has a clear indication of a stroke, he/she should be directly transferred to the stroke team for evaluation and treatment. For non-obvious cases, \textit{DeepStroke} will be applied for the reference of the triage team and ER doctors to reduce misdiagnosis.}

To evaluate \textit{DeepStroke}, we construct a stroke patient video/audio dataset that records the facial motions of the patients during their process of performing a set of vocal speech tests when they visit the ER. The recruited participants are all showing some level of neurological conditions with suspicion of stroke when visiting the ER. This is closer to the real ER scenarios and {\it much more challenging} than distinguishing stroke patients from healthy people. 
Our dataset includes diverse patients of different genders, races/ethnicity, ages, and at different levels of stroke conditions; the subjects are free of motion constraints and are in arbitrary body positions, illumination conditions, and background scenarios, which can be regarded as ``in-the-wild.'' 
Experiments on our dataset show that the  \textit{DeepStroke} framework achieves higher performance than the typical stroke triage team and even outperform trained ER clinicians for stroke (who has more clinical information available) while maintaining a manageable computation workload.

A preliminary version of this work was presented in our earlier publication~\citep{MICCAI}. In this paper, we substantially extend our prior work in the following aspects. First, we substitute the text transcript inputs with the spectrogram input and replace the text LSTM with ResNet-18 to maintain the most of speech information, reduce errors induced by the transcription process as well as ease the later feature fusion with video module. Secondly, we introduce a lateral connection to the two branches to resolve the unstable convergence of the network caused by different training dynamics of branches and share low-/high-level features for a better combination of the global context (audio) and local representation (video). {Thirdly, we adopt a transfer learning pipeline and introduce state-of-the-art fairness-aware face image pre-training and image classification pre-training to reduce facial feature biases, mitigate network overfitting and improve network generalizability.} Fourthly, we introduce an adversarial training scheme aiming to remove the subject identity features from the input to alleviate the ``subject-remembering'' effect of the deep neural networks. Lastly, we significantly enlarge the dataset with newly collected data to support extensive new cross-validation experiments and hold-out study on our proposed new framework. 

The {\bf main contributions} of this work are summarized below.
\begin{enumerate}
    \item We constructed a real ER patient facial video and vocal audio dataset for stroke screening with diverse participants. The videos are collected ``in the wild,'' with unconstrained patient body positions, environment illumination conditions, or background scenarios. 
    \item We propose a deep-learning multimodal framework, \textit{DeepStroke}, that highlights its video-audio multi-level feature fusion scheme that combines global context to local representation, the adversarial training that extracts identity-free stroke features, {the transfer learning that improves generalizability}, and the spatiotemporal proposal mechanism for frontal human facial motion sequences. 
    \item The proposed multi-modal method achieves high diagnostic performance and efficiency on our proposed challenging dataset and outperforms the clinicians, demonstrating its high clinical value to be deployed for real ER use. 
\end{enumerate}

\section{Related Work}
\subsection{Facial Motion Analysis for Medical Diagnosis}
In the past two decades, researchers have been striving for more and more novel facial motion analysis methods to achieve accurate detection and evaluation of facial paralysis and other medical conditions. Early work focused on static image analysis. \cite{toxinA} aimed to detect the improvement in abnormal facial movements after treatment with botulinum toxin A; \cite{Dong2008} introduced a ``face nerve index'' based on coordinates of image feature points and achieved an accurate classification of facial paralysis levels. Some others utilized multiple images in a sequence for improved analysis, {\it e.g.},~\cite{He2007} and~\cite{biometrics2004}. With video data, researchers can capture between-frame motions of patient faces and temporal patterns in facial abnormalities. For example,~\cite{Frey2008} analyzed the video trajectories of facial regions for face paralysis detection and evaluation;~\cite{parkinsonvideo} extracted articulatory movements of patient faces in a video to identify patterns of Parkinson's disease. A small video dataset for facial paralysis has been offered by~\cite{MEEI} along with the eFACE~\citep{eface}, House-Brackmann~\citep{hb}, and Sunnybrook~\citep{sunny} metrics. There is not a publicly-available image or video database specifically for the evaluation of stroke within a patient due to the concern of personal identity leakage and IRB constraints; if no real patient case is available, synthetic datasets like CK+~\citep{ckp} have to be used, where only healthy participants are involved. Biases in genders, ages, and races will arise as the majority of the prior datasets only have a limited number of subjects. 

With recent developments in computer vision and especially face alignment, the facial landmark is becoming a popular representation of facial patterns. \cite{Song2017fnp} measured asymmetry for facial movement images based on facial landmarks; \cite{isbi2018} and~\cite{2dlandmark2019} also introduced similar facial landmark-based methods on video or image data. \cite{wang2016} and~\cite{BHI2019} stratified the face of a patient into multiple regions and considered regional information with landmarks. While easy to obtain, the landmarks are not robust and suffer greatly from illumination, occlusion, and motion. Some more recent work designs deep learning methods for the task. \cite{song2018} developed an Inception-DeepID-FNP neural network for paralysis classification, \cite{umcnn2018} considered a multi-task network for both face detection and facial asymmetry evaluation, and \cite{isbi2017} introduced a deep learning framework to classify facial paralysis with regard to the H-B standard. However, all these deep learning frameworks suffer from non-standard head poses or non-facial motions. To address the alignment issues, the majority of the prior work sets up experimental settings with uniform illuminations and puts up hard constraints on the subjects' heads to avoid alignment issues. Such simplification of scenarios greatly hinders the clinical value of these proposed methods, especially in an emergency setting.

Upon reviewing the above literature, we concluded that existing approaches either evaluate their methods between subjects that are normal versus those with clear signs of a stroke, deal with only synthetic data, fail in capturing the spatiotemporal details of facial muscular motions, or rely on experimental settings with hard constraints. Without considering challenges in real emergencies such as illumination variations, pose changes, skin color, and subtleties of patterns, such methods are not practical in real clinical practice or for patient self-assessment.

\subsection{Disentangle Representation Learning}
Our work is also closely related to disentangle representation learning. \citet{tran2017disentangled} developed DR-GAN to learn a generative and discriminative representation for pose-invariant face recognition. \citet{liu2018exploring} proposed the $D^2AE$ framework to adversarially learn the identity-free features for identity verification and the identity-dispelled features to fool the verification system. \citet{zhang2019gait} proposed an AutoEncoder framework to explicitly disentangle pose and appearance features from RGB imagery and the LSTM-based integration of pose features over time produces the gait feature. \citet{lu2019unsupervised} developed an unsupervised domain-specific image deblurring framework by disentangling the content and blur features. \citet{lee2018diverse} proposed an image-to-image translation framework by disentangling image features into a domain-invariant content space and a domain-specific attribute space and combining them to produce diverse output. 

Especially, learning identity-free stroke patterns for diagnosis is a similar process to extracting identity-free features for facial expression recognition (FER), which also aims to detect patterns of facial movements. \citet{7961791} designed identity-contrasting loss that goes in parallel with the expression-related loss to achieve identity-invariant expression recognition. \citet{cai2019identity} proposed a conditional generative model to transform an average neutral face into an average expressive face with the same expression as the input image that is naturally identity-free. \citet{robust} constructed a pose discriminator and a subject discriminator to classify the pose and the subject from the extracted feature representations, respectively, to make them robust to poses and subjects. In this work, we adopt a similar approach and set up identity-related discriminators, and use an adversarial training scheme to extract identity-free features. To our knowledge, our work is the first attempt that introduces adversarial training loss to stroke diagnosis tasks. 

\subsection{Multimodal Deep Learning}
Deep networks that learn features over multiple modalities via shared representation learning~\citep{NgiamKKNLN11} have achieved outstanding performance on various multimedia tasks. Intuitively, when more than one form of data is available, a deep network that utilizes most of them and takes aligned information from different sources, where each modality tends to complement each other and incorporate richer information, will usually result in higher task performance than using these information sources individually. The idea of multimodal deep learning has been making great improvements to domains including emotion recognition~\citep{kahou2016emonets}, disease diagnosis~\citep{xu2016multimodal}, object detection~\citep{7353446}, etc. Common modalities include text, audio, and video, and they complement each other within a multi-modal neural network. 

Specifically for disease diagnosis, multimodal methods have been proven useful to integrate patient data from multiple platforms or in different forms.  \citet{10.1007/978-3-319-75238-9_13} proposed a multi-view deep learning framework (MvNet) with three branches to segment multimodal brain images from different view-points, \textit{i.e.} slices along x-, y-, z-axes; \citet{10.1007/978-3-030-24097-4_1} developed a multimodal deep machine learning architecture for hepatocarcinoma diagnosis with computed tomography images, laboratory test results, anthropometric and sociodemographic data as input; to model the difficulties to start or to stop movements for patients with Parkinson's disease, \citet{8444654} conducted multimodal deep learning over information from speech, handwriting, and gait; \citet{6977954} proposed multimodal deep belief network (DBN) to cluster cancer patients observation data from different platforms. The fusion of medical imaging and electronic health records using deep learning has been another heated topic in recent years~\citep{huang2020fusion}, which will make pixel-based models and contextual data from electronic health records (EHR) work cooperatively for the diagnostic purpose. 

Stroke diagnosis can also benefit from multimodal deep learning. To diagnose a stroke case, neurologists detect the facial deficits on the patient's face or the disorder in speech---the two modalities tend to work jointly. In this work, we adopt the multimodal idea to allow the facial video and vocal audio to work together for a final diagnosis.

\section{Dataset}
The clinical dataset for this study was acquired in the ERs of the Houston Methodist Hospital in Texas by the physicians and caregivers from the Eddy Scurlock Stroke Center at the Hospital under an IRB-approved study.\footnote{{This study is conducted under Houston Methodist IRB protocol No. Pro00020577, Penn State IRB site No. SITE00000562.}} It has taken more than a year for us to recruit a large pool of patients in various stroke-related emergencies, and the cohort is still expanding. The subjects enrolled are patients with suspicion of stroke 
while visiting the ER while obvious stroke cases with severe symptoms are excluded.  To help preserve the patients' personal information, we only transmitted limited information that we think is sufficient for this study between institutes to avoid any identifiable information to be collected and dispensed. The patients are only assessed when they're in a relatively stable condition so the collection of data will not impose extra risk on high-emergency cases. Note that the gender ratio of our dataset is relatively balanced without intervention, and the race/ethnicity or age distributions are not manually controlled, which would roughly represent the real distribution of the incoming ER patients. Potential bias will be discussed in Sec.~\ref{Sec: experiments}.

\begin{figure}[ht]
\begin{subfigure}{.25\textwidth}
  \centering
  \includegraphics[width=.95\linewidth]{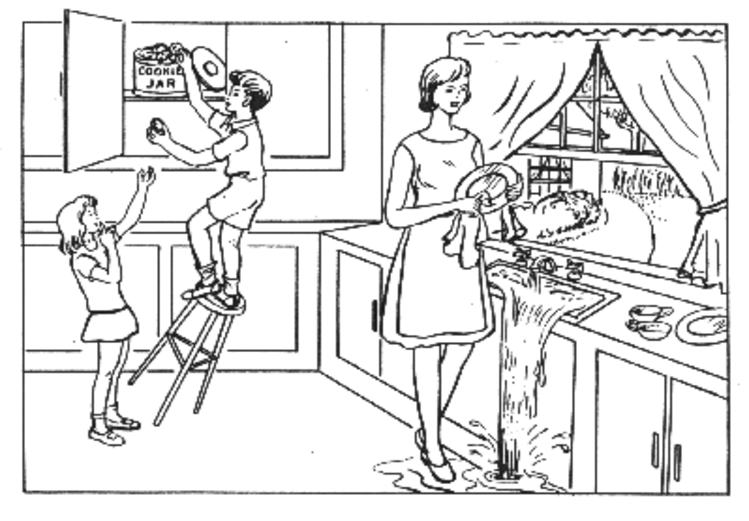}  
  \caption{The ``Cookie Theft'' picture}
  \label{fig:CTP}
\end{subfigure}
\begin{subfigure}{.22\textwidth}

  \centering
  \includegraphics[width=.95\linewidth]{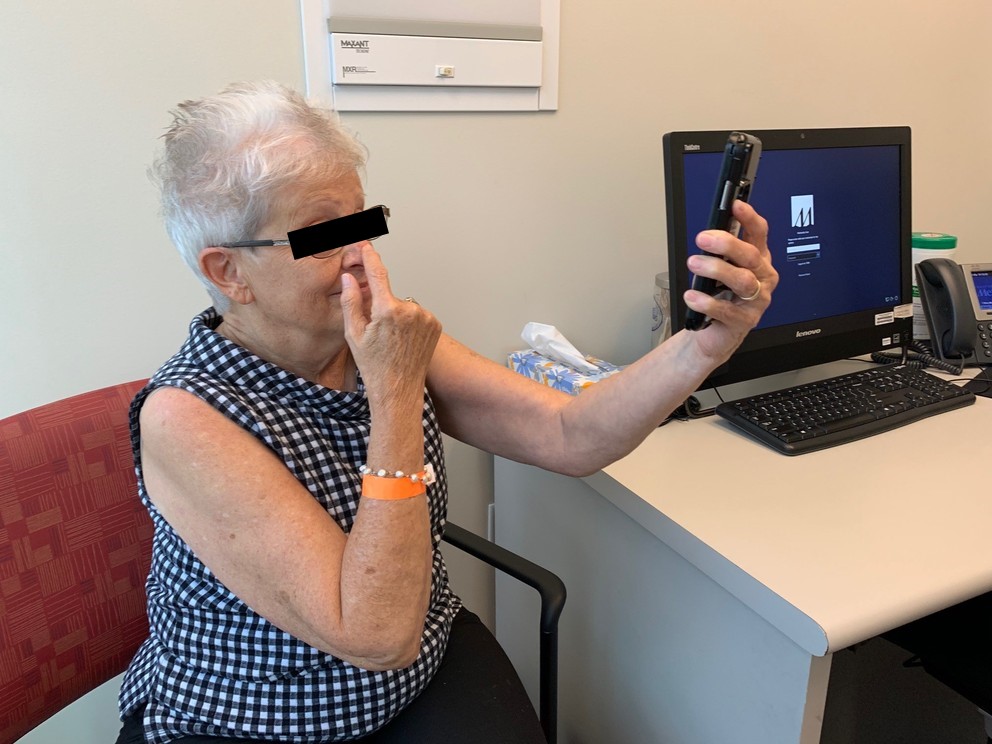}  
  \caption{{Setup of the recording protocol}}
  \label{fig:setup}
\end{subfigure}
\caption{The speech task and the sample recording protocol.}
\label{fig:set}
\end{figure}

Clinically, the ability of speech is an important and efficient indicator of the presence of stroke and is the preferable measurement doctors will use to make initial clinical impressions; if a potential patient slurs, mumbles, or even fails to speak, he or she will have a very high chance of stroke~\citep{harbison2003diagnostic,kothari1999cincinnati}. During the evaluation and recording stage, we follow the NIH Stroke Scale~\citeyearpar{nihss} and perform the following speech tasks on each subject: (1) We ask the patient to repeat the sentence ``it is nice to see people from my hometown,'' and (2) we ask the patient to describe the ``Cookie Theft'' picture as shown in Figure.~\ref{fig:CTP}. 

The ``Cookie Theft'' task requires the subject to retrieve, think, organize, and express the information, which will evaluate the patient's speech ability from both motor and cognitive aspects. It has been making great success in identifying patients with Alzheimer's-related dementia, aphasia, and some other cognitive-communication impairments~\citep{giles1996performance}, and would be suitable for our stroke screening purpose.

The subjects are video recorded when they are performing the two tasks. The collection protocol is set up with an Apple iPhone X, as shown in Figure~\ref{fig:setup}. Note that a phone rack and an auxiliary microphone will be used if the patient is too weak to hold the device stably. Each video is accompanied by metadata information on both clinical impressions by the ER physician (indicating the doctor's initial judgment on whether the patient has a stroke or not from his/her speech and facial muscular conditions) and ground truth from the diffusion-weighted MRI (including the presence of acute ischemic stroke, transient ischemic attack (TIA), etc.). 
Samples of such metadata information are shown in Table~\ref{tab:meta}. 

\begin{table}[htbp]
\scriptsize
\setlength\tabcolsep{1.5pt}
\renewcommand\arraystretch{.8}
\caption{Metadata info for subjects enrolled. (PE: Patient Example \#)}\label{tab:meta}
\resizebox{\linewidth}{!}{
\begin{tabular}{lllll}
\toprule
\textbf{PE}   & \textbf{Complaint}                                                       & \textbf{Clinical Impression}                                                                       & \textbf{Discharge Diagnoses}                                                                                                                                                                                                                                            & \textbf{Details}                                                                                                                                                                               \\ \midrule
1 & Weakness                                                                 & DR: Stroke / TIA                                                                          & \begin{tabular}[c]{@{}l@{}}1, Acute ischemic stroke\\ 2, Dyslipidemia \\ 3, Type 2 diabetes A1c\end{tabular}                                                                                                                                                        & \begin{tabular}[c]{@{}l@{}}L cortical infarct (precentral \\ gyrus) and additional R \\ corpus callosum (subacute)\end{tabular}                                                       \\ \midrule
2 & \begin{tabular}[c]{@{}l@{}}Other - \\ Neurologic \\ problem\end{tabular} & DR: Stroke / TIA                                                                          & \begin{tabular}[c]{@{}l@{}}1, TIA\\ 2, possible TIA versus\\     seizure\\ 3, acute metabolic \\     encephalopathy\\ 4, UTI\\ 5, Obesity\\ 6, HLD\\ 7, HTN\\ 8, Pre-diabetes\end{tabular} & \begin{tabular}[c]{@{}l@{}}right vertebral artery, V3, V4\\ Unable to obtain MRI due to \\ metal artifact\end{tabular}                                                                \\ \midrule
3 & \begin{tabular}[c]{@{}l@{}}Tremor, \\ Stuttering\end{tabular}     & \begin{tabular}[c]{@{}l@{}}Aphasia, Ataxia, \\ Unspecified Tremor, \\ CVA\end{tabular}    & 1, Tremors                                                                                                                                                                                                                                                          & \begin{tabular}[c]{@{}l@{}}Neurology gave pt DDx of acute\\ ischemic stroke, TIA, and toxic-\\ -metabolic. Workup ruled out \\ CVA and indicated breakthrough\end{tabular} \\ \midrule
4 & ...                                                                   & ...                                                                                       & ...                                                                                                                                                                                                                                                            & ...                                                                                                                                                                                   \\ \midrule
\end{tabular}
}
\end{table}
{Note that in this dataset, the Clinical Impression is given after the ER screening by the doctors who usually have access to emergency imaging reports, vitals, and other information in the Electronic Health Records (EHR). Some early-enrolled patients are not included later in this study due to missing clinical impression data resulting from being transferred from another healthcare provider. }

We have been recruiting new subjects to expand the dataset in the last few years. Up to the time of completing the manuscript, 108 males and 113 females have been recruited, non-specific of age, race/ethnicity, or the seriousness of stroke. Among the 221 individuals, 159 are patients diagnosed with stroke using MRI, 62 are patients who do not have a stroke but are diagnosed with other clinical conditions. A summary of demographic information is shown in Table~\ref{tab:demo}. In this work, we formulate the diagnosing process as a binary classification task and only attempt to identify stroke/TIA cases from non-stroke cases. Though there are varieties of stroke subtypes, binary output has been sufficient to function as a screening decision in ER. 

\begin{table}[htbp]
\footnotesize
\centering
\setlength{\tabcolsep}{7.5pt}
\caption{{Summary of subjects' demographic information in the dataset.} }\label{tab:demo}
\begin{tabular}{ccccc}
\toprule
  \textbf{Attribute}                &    \textbf{Group}         & \textbf{Stroke} & \textbf{Non-Stroke} & \textbf{Total} \\\midrule
\multirow{2}{*}{Gender}    & Male             &    76    &     32       &   108    \\
                           & Female           &    83    &     30       &   113    \\\midrule
\multirow{2}{*}{Age}       & $\leq$ 65 y/o    &    80    &     36       &   116    \\
                           & $\geq$ 65 y/o    &    79    &     26       &   105    \\\midrule
\multirow{3}{*}{Ethnicity} & Hispanic         &    12    &      4       &   16    \\
                           & Non-Hispanic     &    146   &     52       &   198   \\
                           & Opt-out          &    1     &     6        &    7    \\\midrule
\multirow{2}{*}{Race}      & African American &    54    &     18       &   72    \\
                           & Other            &    105   &     44       &   149   \\\midrule
\multicolumn{2}{c}{All subjects}              &    159   &     62       &   221    \\
\bottomrule
\end{tabular}
\end{table}

Our dataset is unique, as compared to existing ones~\citep{isbi2017,He2007}, because our cohort consists of actual patients visiting the ERs and the videos are collected under unconstrained, or ``in-the-wild'' conditions. Existing work generally setup experimental settings before collecting the image or video data, which will result in uniform illumination conditions and minimum background noise. In our dataset, the patients can be in bed, sitting, or standing, where the background and illumination are usually not under ideal control conditions. Furthermore, previous work often enforced rigid constraints over the subjects' head motions, which sidesteps the alignment challenges and makes the unrealistic assumption of having stable face poses. We only ask patients to focus on the instructions, without rigidly restricting their motions. We use intelligent video-processing methods to accommodate for the ``in-the-wild'' conditions. The acquisition of facial data in natural settings allows comprehensive evaluation of the robustness and practicability of our work for real-world clinical use, remote diagnosis, and self-assessment in most settings.

\section{Methods} 
\begin{figure*}[htbp]
    \centering
    \includegraphics[page=6,width=\textwidth,trim={0cm 10.4cm 12.3cm 0cm}]{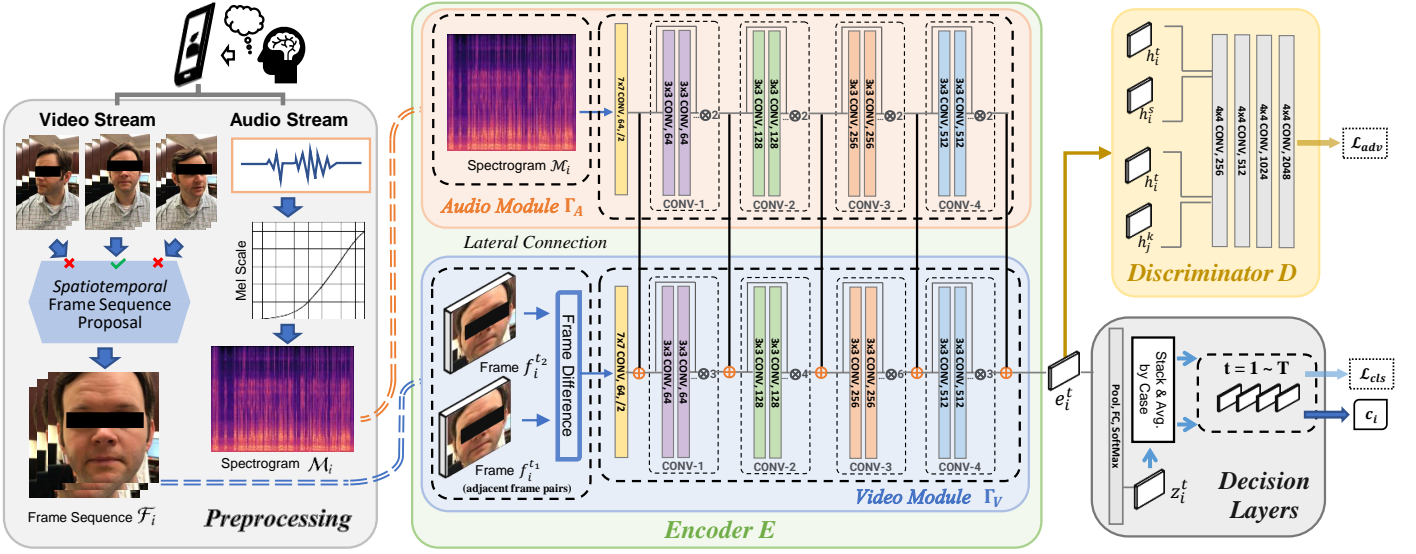}
    \caption{The training framework of \textit{DeepStroke}. The preprocessing part involves a spatiotemporal facial frame sequence proposal and a transformation of audio features to spectrograms. In the encoder $E$, at any timestamp $t$, the input frame pairs $f_i^{t_1}$ and $f_i^{t_2}$ are guaranteed to be adjacent pairs, and the total spectrogram is duplicated and appended as the audio features for time $t$. The symbol $\oplus$ indicates the concatenation with the lateral connection.  }
    \label{fig:model}
\end{figure*}
Fig.~\ref{fig:model} shows the training framework of \textit{DeepStroke}. 
We first preprocess the $i$-th input raw video to obtain the facial-motion-only, near-frontal face sequence $\mathcal{F}_i$ and its corresponding audio spectrogram $\mathcal{M}_i$. Then we extract the audio-visual feature $e_i$ from $\mathcal{F}_i$ and $\mathcal{M}_i$ by a lateral-connected dual-branch Encoder $E$, which includes a video module $\Gamma_{\text{v}}$ for local visual pattern recognition, and an audio module $\Gamma_{\text{a}}$ for global audio feature analysis. A subject Discriminator $D$ is also employed to help $E$ learn features that are insensitive to subject identity difference but are sensitive to distinguishing stroke from non-stroke. 
When training, we use the case-level label as a pseudo label for each video frame and train the \textit{DeepStroke} network as a frame-level binary classification model. We also sample the intermediate output feature maps from different videos to train $D$ and $E$ adversarially.
During inference, we first perform frame-level classification and then calculate the case-level predictions by averaging over all frames' probabilities to mitigate frame-level prediction noise. 
More details are described as follows.

\subsection{Data Preprocessing}
For each raw video, we propose a spatiotemporal proposal mechanism to extract frontal-face sequences from the raw video.  For each audio extracted from the raw video, we transform the soundtrack into a spectrogram that represents the amplitude at each frequency level over time. 

\textbf{Spatiotemporal proposal of facial action video:} In facial motion analysis, one challenge is to achieve good face alignment. As our data are collected ``in the wild'', we introduce a pipeline to extract frame sequences with near-frontal facial pose and minimum non-facial motions. First, we detect and track the patient's face with a rigid, square bounding box and estimate the poses. Frame sequences 1) estimated with significant roll, yaw, or pitch, 2) showing continuously changing pose metrics, or 3) having excessive head translation or too little change estimated with optical flow magnitude with the previous frame, are excluded (detailed criteria are presented in Section~\ref{S5.1}). A video stabilizer with a sliding window over the trajectory of between-frame affine transformations smooths out pixel-level vibrations on the sequences before they are passed to the Encoder $E$.

\textbf{Speech spectrum analysis:} 
Instead of transcribing the vocal audio to text corpus \citep{MICCAI}, which may suffer from translation errors,
we turned to spectrograms for speech analysis due to the following reasons: (1) Spectrogram is the complete representation of an audio file since the amplitude over frequency bands are captured over time. Recent deep learning-based transcription models commonly transform soundtracks into spectrograms for their classification models~\citep{hannun2014deep,amodei2016deep}. (2) By choosing spectrogram, which is image-like input, we can adapt similar networks for the two branches and ensure they have rather similar training dynamics to converge at a similar pace, which will make them cooperate better. In this work, we use the Mel Scale~\citep{stevens1937scale} and Fast Fourier Transform (FFT) to transform the audio signal before plotting the spectrogram.

\subsection{Model Design}
As Fig.~\ref{fig:model} shows, after preprocessing the raw input videos, 
we further extract stroke-discriminative and identity-free features from the input video and audio via the feature encoder $E$ and the subject discriminator $D$. 

\textbf{Feature encoder:} 
Let $\mathcal{F}_i=(f^1_i,\cdots, f^T_i)$ denotes a sequence of $T$ temporally-ordered frames from the $i$-th input video and its corresponding spectrogram is $\mathcal{M}_i$. We extract their features through feature encoder $E$, which includes one video module $\Gamma_\text{v}$ and one audio module $\Gamma_\text{a}$, fused by lateral connection.

\textit{$\triangleright$ Video Module.}
To extract temporal visual features from input $\mathcal{F}_i$, a pair of adjacent frames from $\mathcal{F}_i$ is forwarded to the video module $\Gamma_{\text{v}}$. Due to the frame-proposal process, the original frame sequence sometimes has long gaps between two nearby frames, which will result in large, non-facial differences being captured. We tackle this by keeping track of the frame index and only sample a specific number of real adjacent frame pairs in $\mathcal{F}_i$ (frame $f_i^{t_1}$ and $f_i^{t_2}$) to extract local visual information. 
Instead of directly inputting a pair of frames, for better capturing subtle facial motions between adjacent frames, we compute the image difference between $f_i^{t_1}$ and $f_i^{t_2}$ and then pass it through the network as the feature for the frame pair $f^t_i$.

\textit{$\triangleright$ Audio Module.}
To extract disease patterns from the input audio spectrogram $\mathcal{M}_i$, we feed $\mathcal{M}_i$ to the audio module $\Gamma_{\text{a}}$. Since $\mathcal{M}_i$ contains the whole temporal dynamics of the input audio sequence, we append $\mathcal{M}_i$ to each frame pair $x^t_i$ and $x^{t+1}_i$ to provide a global context for the frame-level stroke classification.

\textit{$\triangleright$ Lateral Connection.}
To effectively combine features of video $\mathcal{F}_i$ and audio $\mathcal{M}_i$ at different levels, we also introduce lateral connection \citep{xiao2020audiovisual} between the convolutional blocks of the video module $\Gamma_{\text{v}}$ and the audio module $\Gamma_{\text{a}}$. To ensure the features are aligned when being appended, we perform $1\times 1$ convolution \citep{lin2013network} to project the global audio feature to the same embedding space as the local frame features and then sum them up. Compared with the late fusion of two branches used in our prior work \citep{MICCAI},
lateral connections-based fusion not only combines more levels of features but also enables different branch dynamics to stay similar, which will maintain the convergence rate of each branch to be relatively closer and help the global context better complement the local context during the training stage. 

\textbf{Subject discriminator:}
Due to the relatively small number of available videos, it is easy and tempting for the encoder $E$ to memorize the facial and audio features of each subject and just match testing subjects with training subjects based on similarity in appearance and voice when performing the inference. To avoid this issue of classification based on subject-dependent features, we further design a subject discriminator $D$ with an adversarial learning idea to help encoder $E$ learn identity-free features. The discriminator $D$ is designed to simply distinguish whether the input pair of intermediate features from encoder $E$ are from the same subject or not and will be used to adversarially train the encoder $E$. {In the implementation, when we feed the (original) input data to the network, we will take another data batch (denoted as the adversarial data) from either the same case or a different case with equal probability. When training $D$, both original and adversarial data will be processed by the encoder $E$ with parameters frozen for feature embeddings. If the original and adversarial data came from the same case, we assign a positive label to the embedding pair (and a negative label otherwise). The generated feature embedding will include both information from the audio and video (thanks to the lateral connection in the encoder $E$) and are used to train $D$. An adversarial loss is given to update $D$. The adversarial loss is also added to the classification loss for the later update of encoder $E$.} Details of the loss calculation is introduced in Sec.~\ref{sec:loss}
\begin{figure}[htbp]
    \centering
    \includegraphics[page=9,width=\textwidth,trim={0.4cm 12.5cm 12.8cm 0cm}]{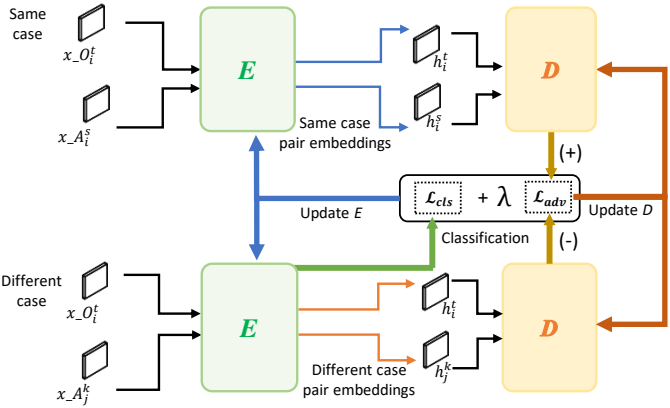}
    \caption{{Workflow of the discriminator $D$ of \textit{DeepStroke}. $\mathcal{L}_{\text{cls}}$ will be calculated first using $x\_O_i^t$, the original input (including $f_i^{t_1}$ and $\mathcal{M}_i$), and generate feature embedding $h_i^t$. Then the encoder will freeze parameters and generate $h_i^s$ and $h_j^k$ for $x\_A_i^s$ and $x\_A_j^k$, the adversarial inputs from the same/different cases. Positive label $(+)$ is assigned to same-case pairs and negative $(-)$ to different-case pairs. $\mathcal{L}_{\text{adv}}$ is then calculated to update $D$, and a weighted sum of $\mathcal{L}_{\text{cls}}$ and $\mathcal{L}_{\text{adv}}$ is used to update $E$. }}
    \label{fig:adv}
\end{figure}

In theory, the network of \textit{DeepStroke} can employ various networks as its backbone. Here we choose ResNet-34 for $\Gamma_{\text{v}}$ and ResNet-18 for $\Gamma_{\text{a}}$ to accommodate the relatively small size of the video dataset and the simplicity of the spectrogram, and to reduce the computational cost of the framework. For the discriminator $D$, we follow the design of DCGAN \citep{radford2015unsupervised} with four convolution layers and binary output. 
For the intermediate feature input to the discriminator, {we take the frame-level feature $e_i^t$ (output feature from the decoder) for case $i$ at time $t$. The choice of intermediate feature is ablated in Section~\ref{S5.3.1}.}

\textbf{Decision layers:}
During inference, because our model is trained with pseudo frame-level labels, to mitigate frame-level prediction noise, we first perform frame-level classification using encoder $E$ into final fused features $e^t_i$. $e^t_i$ is then passed through the fully-connected and softmax layers to generate the frame-level class probability score $z_i^t$. The case-level prediction $c_i$ is then obtained by stacking and averaging the frame-level predictions (stroke probability ranging from 0 to 1) and compare with the decision threshold. 

\subsection{Model Training}
\subsubsection{Training with Transfer Learning}
{Due to the relatively small number of samples available, the proposed deep neural network can overfit on the training samples and result in a ``face-remembering'' effect. Besides employing the subject discriminator to alleviate this effect, we adopt a transfer-learning framework to train \textit{DeepStroke} by starting with pre-trained network backbones and freezing the parameters.
Only the final output layers and the discriminator module are fine-tuned. For the video module network, we employ state-of-the-art fairness-aware face image pre-training, the FairFace~\citep{karkkainen2021fairface} ResNet-34 model 
that was pre-trained on a face image dataset that has evenly distributed race, age, and ethnicity attributes. This aims to reduce the common facial-attribute biases and prevent network overfitting on irrelevant features. For the audio module, we apply an ImageNet-pre-trained ResNet-18 backbone as the transfer learning starting point to mitigate overfitting. The transfer learning improves the generalizability of the network and will be demonstrated in Sec.~\ref{sec:abl}.
}
\subsubsection{Loss Functions}~\label{sec:loss}
\textbf{Classification loss}: To help $E$ learn stroke-discriminative features, we use a standard binary cross-entropy loss between the prediction $z^t_i$ and video label $\mathcal{Y}_i$ for all the training videos and their $T$ frames:
\begin{equation}
    \label{l_cls}
    \mathcal{L}_{\text{cls}}(E) = -\sum_{i}\sum_{t}\left(\mathcal{Y}_i\log z^t_{i} + (1-\mathcal{Y}_i)(1-\log z^t_{i})\right)\enspace.
\end{equation}

\textbf{Adversarial loss:} To encourage Encoder $E$ to learn identity-free features, we introduce a novel adversarial loss to ensure that the output feature map $h^t_i$ does not carry any subject-related information. We impose this via an adversarial framework between the subject discriminator $D$ and feature encoder $E$, as shown in Fig. \ref{fig:model}. The latter, $E$, will provide a pair of feature maps, either computed from the same subject video $(h^t_i, h^s_i)$ at time $t$ and $s$, or from different subject videos $(h^t_i, h^k_j)$ at time $t$ and $k$, where time $s$ and $k$ can be randomly chosen. Discriminator $D$ then attempts to classify the pair as being from the same/different subject video using an $l_2$ loss, as LS-GAN \citep{mao2017least} adopts:
\begin{equation}
    \label{l_adv_d}
    \mathcal{L}_{\text{adv}}(D) = -\sum_{i}\sum_{t}\left(\left\|D(h^t_i, h^s_i)\right\|_2 + \left\|1- D(h^t_i, h^k_j)\right\|_2\right)\enspace.
\end{equation}
The adversarial framework further imposes a loss function on the feature encoder $E$ that tries to maximize the uncertainty of the discriminator $D$ output on the pair of frames:
\begin{equation}
    \label{l_adv_e}
    \mathcal{L}_{\text{adv}}(E) = -\sum_{i}\sum_{t}\left(\left\|\frac{1}{2}-D(h^t_i, h^s_i)\right\|_2 + \left\|\frac{1}{2}- D(h^t_i, h^k_j)\right\|_2\right)\enspace.
\end{equation}
Thus the encoder $E$ is encouraged to produce features that the discriminator $D$ is unable to classify if they come from the same subject or not. In so doing, the features $h$ cannot carry information about subject identity, thus avoiding the model to perform inference based on subject-dependent appearance/voice features. Note that our model is different from classic adversarial training used in GANs \citep{goodfellow2014generative} because we only focus on classification and there is no generator network in our framework.

\textbf{Overall training objective}:
During training, we minimize the sum of the above losses:
\begin{equation}
    \label{eq:l_total}
    l = \mathcal{L}_{\text{cls}}(E) + \lambda (\mathcal{L}_{\text{adv}}(E)+\mathcal{L}_{\text{adv}}(D))\enspace,
\end{equation}
where $\lambda$ is the balancing parameter. The first two terms can be jointly optimized, but the discriminator $D$ is updated while the encoder $E$ is held constant.


\section{Experiment}\label{Sec: experiments}
To better present the details of our proposed method, we first introduce the setup and implementation details and then show the comparative study with baselines. We also ablate the power of model components and structures to validate our design. 

\subsection{Setup and Implementation}\label{S5.1}
The whole framework is running on \texttt{Python} 3.7 with \texttt{Pytorch} 1.1, \texttt{OpenCV} 3.4, \texttt{CUDA} 9.0, and \texttt{Dlib} 19. Both ResNet models are pre-trained on ImageNet~\citep{imagenet}. Each .mov file from the iPhone will first be separated as frame sequences as batches of .png files and one global .wav audio file.

For the frame sequences, we detect the location of the patient's face as a square bounding box with \texttt{Dlib}'s face detector~\citep{dlib09} and track it using the \texttt{Python} implementation of the ECO tracker~\citep{danelljan2017eco}. We perform pose estimation by solving a direct linear transformation from the 2D landmarks predicted by \texttt{Dlib} to a 3D ground truth facial landmarks of an average face, which resulted in three values corresponding to the angular magnitudes over three axes---pitch, roll, and yaw. To tolerate estimation errors, we regard those with angular motions less than a threshold $\beta_1$ as frontal faces. A 5-frame sliding window records the between-frame changes in the pose. If the total changes (three axes) sum up to more than a threshold $\beta_2$, we abandon the frames starting from the first position of the sliding window. In the meantime, the between-frame changes are measured by optical flow magnitude. If the total estimated change is smaller than $\beta_l$ (no motion) or larger than $\beta_h$ (non-facial motions), we also exclude the frame. We empirically set $\beta_1=5^\circ$, $\beta_2=20^\circ$, $\beta_l = 0.01$, and $\beta_h = 150$. After manipulation, we crop the size of each frame to $224\times224\times3$ to align with the ImageNet dataset. The real frame numbers are kept to ensure that only adjacent frame pairs are loaded to the network. 

For the audio files, we use \texttt{librosa} to load and trim the soundwave, and plot the Log-Mel spectrogram, where the horizontal axis is the temporal line, the vertical axis is the frequency bands, and each pixel shows the amplitude of the soundwave at the specific frequency and time. The output spectrogram is also set to the size of $224\times224\times3$.

The entire pipeline is trained on a server computer with a hex-core CPU, 64GB RAM, and an NVIDIA GPU with 24GB VRAM. To accommodate for the class imbalance inside the dataset and ER setting, a higher class weight of 2.0 is assigned to the non-stroke class. {The default learning rate is set to 1e-7 with decay after every five epochs, and we early stop at epoch 20 due to the quick convergence of the network. The batch size is set to be 64 and $\lambda$ in \eqref{eq:l_total} is set to be 10 for loss scaling purposes. Model parameters and learning rates are then tuned separately for each model/baseline in each experiment. 
} 

\subsection{Baselines and Experiments}
\begin{table*}[htbp]
\small
\centering
\renewcommand\arraystretch{1.2}
\setlength{\tabcolsep}{14pt}
\caption{Results of the cross-validation experiment. Raw: results with decision threshold 0.5 after the final softmax output; aligned: results with the threshold that makes the specificity aligned {with the Triage performance. Due to the dataset limitation, exact alignment is not possible. We take and report the closest possible alignment value}. The best performance of each metrics among the model/baselines is highlighted in bold.}
\begin{tabular}{rccccccc}
\toprule
\multicolumn{1}{r}{\multirow{2}{*}{}} & \multicolumn{2}{c}{\textbf{Accuracy (\%)}}                                     & \multicolumn{2}{c}{\textbf{Specificity (\%)}}          & \multicolumn{2}{c}{\textbf{Sensitivity (\%)}}                                  & \textbf{AUC}                             \\ \cline{2-8} 
\multicolumn{1}{c}{\textbf{Model/Baseline}}                                                              & Raw                             & Aligned                         & Raw                             & Aligned & Raw                             & Aligned                         & ---                             \\ \midrule
Dummy                                           &  \multicolumn{2}{c}{71.95}   &   \multicolumn{2}{c}{0.00}   &  \multicolumn{2}{c}{100.00}  &       0.5000                    \\ \midrule
Module $\Gamma_\text{a}$ (Audio)                &       62.89        &      63.35         &       24.19        &      45.16         &       77.99        &     70.44          &          0.5998                 \\ 
Module $\Gamma_\text{v}$ (Video)                &       68.32        &      62.90         &       16.13        &      45.16         &       88.68        &     69.81          &          0.6042                 \\ 
I3D (Video)                                     &       71.04        &      51.13         &       17.74        &      45.16         &       91.82        &     53.46          &          0.5125                 \\ 
SlowFast (Video)                                &       70.13        &      46.61         &       8.01         &      45.16         &   \textbf{94.34}   &     47.17          &          0.4685                 \\ 
MMDL (Audio + Video)                            &       63.80        &      65.16         &       30.65        &      45.16         &       76.73        &     72.96          &          0.6414                 \\ 
\textbf{\textit{DeepStroke}} (Audio + Video)    &  \textbf{72.40}    &   \textbf{71.40}   &  \textbf{32.26}    &      45.16         &       88.05        &   \textbf{81.13}   &      \textbf{0.7163}            \\ \midrule
Triage Stroke Screening (Triage)            & \multicolumn{2}{c}{64.03}                                        & \multicolumn{2}{c}{45.71}                & \multicolumn{2}{c}{70.19}                                        & ---                          \\
Clinical Impression (CI)                                                                & \multicolumn{2}{c}{69.23}                                        & \multicolumn{2}{c}{\textbf{51.61}}        & \multicolumn{2}{c}{76.10}                                        & ---                         \\  
\bottomrule
\end{tabular}\label{tab:baseline}
\end{table*}
We construct baseline models for both video and audio tasks. For each case (video/audio), the ground truth for comparison is the binary diagnosis result obtained through the MRI scan. Our chosen baselines are introduced as follows:
\begin{itemize}
    \item \textbf{Audio module $\Gamma_\text{a}$:} The first corresponding baseline is the strip audio module from the proposed method that takes the spectrograms as input. We use the same setup to train the audio module and obtain binary classification results on the same data splits. The separate audio module takes the same transfer-learning setup as the full model.
    \item \textbf{Video module $\Gamma_\text{v}$:} The other baseline is the strip video module from the proposed method that takes the preprocessed frame sequences as input. We use the same adversarial training scheme and transfer-learning setup to train the video module and obtain binary classification results on the same data splits. 
    \item \textbf{I3D:} The Two-Stream Inflated 3D ConvNet (I3D)~\citep{i3d} expands filters and pooling kernels of 2D image classification ConvNets into 3D to learn spatiotemporal features from videos. It was the state-of-the-art model years ago. For our task, I3D can be inferior because the calculation of optical flow can be time-consuming and result in more noise. 
    \item \textbf{SlowFast:} The SlowFast~\citep{slowfast} network is a video recognition network proposed by Facebook that involves a Slow pathway to capture spatial semantics and a Fast pathway to capture motion at fine temporal resolution. SlowFast achieves strong performance on action recognition in video and has been a powerful state-of-the-art model in recent years. 
    \item \textbf{MMDL:} Our prior work MMDL~\citep{MICCAI} is a preliminary version of the proposed two-branch method that takes similar preprocessed frame sequences for the video branch, but text transcripts for the audio branch. The video branch uses feature difference instead of image difference (which we will ablate later), and the audio branch was an LSTM that performs text classification. Due to drastically different network structures, the two branches only have connections in the final layer using a ``late-fusion'' scheme. 
\end{itemize}


In evaluation, we conducted {both a 5-fold cross-validation experiment and a time-cutoff hold-out experiment. In the cross-validation experiment, the full dataset is split randomly and evenly into 5 folds. Each fold takes four folds for training and the rest for testing, and the model will be reset for each fold. We assess the model performance by the accuracy, specificity, sensitivity, and area under the ROC curve (AUC). In the time-cutoff hold-out experiment, we set the first recruited 168 cases as the training/validation set and the later 53 cases as the testing set. We perform a similar 5-fold cross-validation over the training/validation set and use the saved five top models from each training fold to predict the cases in the testing set. We report the mean, standard deviation, and range of the AUC as both the model performance and stability benchmark. }

{Because our objective is to perform stroke screening for incoming patients, the proposed methods and the baselines are compared to the triage team's performance in stroke pre-hospital screening and the ER doctors' clinical impression we obtained with the metadata. As the real stroke triage in ER often happens when there is stroke onset of the patient, the triage result was not able to be fully collected like the ER doctor's clinical impression in our dataset, due to the high risk of delayed diagnosis and IRB restrictions. The performance measurement of the triage team is from another internal study based on a much larger cohort of patients over a longer period of time. }
\subsubsection{Cross-Validation Experiment}\label{sec:cv5}
{In the cross-validation experiment, 80\% data are used for training and 20\% data are used for testing. We first select one fold as the test set and train a model based on the four other folds. After the experiment using one fold as test set is completed, another folder is used as the test set and a new model is trained using the remaining four folds; note that, the new model is re-initialized when using a new test fold, to ensure the testing data is never seen by the new model. The model that achieves the best result among the first 20 epochs on the current test fold is saved, and the overall performance of the current experiment is calculated with all the five folds' testing results (which covers the whole dataset). Hyperparameters of the experimental setup are tuned based on the overall cross-validation performance on the whole dataset.  }

\begin{figure}[ht!]
    \centering
    \includegraphics[width=.8\linewidth,trim={0.8cm 0.5cm 1.5cm .8cm}]{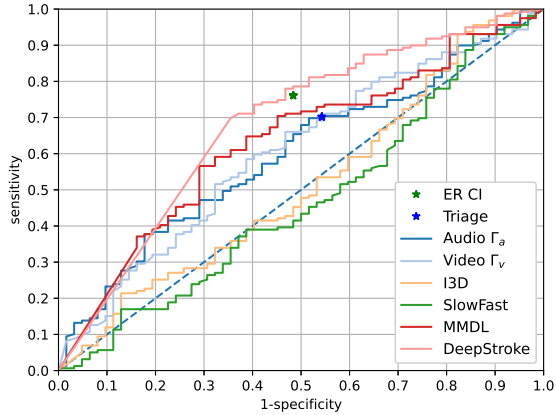}
    \caption{The ROC curve from 5-fold cross-validation for the proposed \textit{DeepStroke} and other baselines. {The blue dot shows the performance of the triage team in stroke screening (Triage), and }the green dot shows the performance of clinical impression (CI) from ER physicians. {The dashed line is for a ``dummy classifier'' that predicts all cases to be stroke/positive.}}
    \label{fig:roc_base}
\end{figure}

{Besides contrasting the model performance with the triage team and ER doctors,} we further examine the effectiveness of our proposed methods {in reducing false negatives and improving stroke screening sensitivity} by aligning the specificity of each method to be the same as the triage performance by changing the threshold for binary cutoffs, while checking and comparing for other measurements. The results are shown in Table~\ref{tab:baseline}. For better comparison, the ROC curves for the proposed model and baselines are also plotted in Fig.~\ref{fig:roc_base}, {together with the triage performance} and clinical impression performance. {For reference, we include the performance of a ``dummy'' classifier that predicts all cases to be stroke/positive.}

From both Table \ref{tab:baseline} and Fig.~\ref{fig:roc_base}, one can see that our proposed \textit{DeepStroke} outperforms several state-of-the-art methods as well as its single module variants. {Both the Audio Module $\Gamma_\text{a}$ and the Video Module $\Gamma_\text{v}$ achieve an AUC level of around 0.6, showing very high sensitivity while suffering from low specificity.} Comparing the Video Module $\Gamma_\text{v}$ with two state-of-the-art video recognition models, we can also see a much higher model performance. When specificity is aligned, Video Module $\Gamma_\text{v}$ achieves 11.77\% higher accuracy, 16.35\% higher sensitivity, and 0.0917 higher AUC than I3D and 16.29\% higher accuracy, 22.64\% higher sensitivity, and 0.1357 higher AUC than SlowFast. {When $\Gamma_\text{a}$ and $\Gamma_\text{v}$ collaborate, a decent performance gain is seen in both our previous work Multimodal Deep Learning (MMDL) and our proposed \textit{DeepStroke}. The MMDL method achieves 0.6414 AUC; when specificity is aligned with Triage performance at 45.16\%, a sensitivity of 72.96\% is reported, which is 2.78\% higher than the Triage performance.} Compared with our prior work MMDL, \textit{DeepStroke} achieves 0.0749 higher AUC. When specificity is aligned to Triage, sensitivity is improved by 8.27\%. This is a major improvement considering the prior work has already demonstrated good performance on the preliminary version of the dataset. 

We believe the improvements came from the following aspects. First, by adopting a different audio representation and introducing the lateral connections, the proposed framework resolves the unstable convergence problem in the prior work caused by different training dynamics of branches, while also sharing low- and high-level features for a better combination of global audio context and local frame features. Secondly, our adversarial training scheme can keep the network from remembering the identity features and extract pure stroke-related features for the network to learn, which also mitigates the overfitting problem. Thirdly, introducing the transfer learning pipeline and adopting fairness-aware face pre-training mitigate the network's overfitting on facial attributes, and improves the generalizability. The three aspects are discussed in the following ablation studies (Sec.~\ref{sec:abl}). Finally, in another aspect, using spectrograms instead of transcripts will maximally preserve the patterns in the original audio files since the soundwave information is fully presented without inference, addition, or deletion.

When compared with the triage team performance, the proposed \textit{DeepStroke} shows 10.94\% higher sensitivity, and 7.37\% higher accuracy while achieving the same specificity, illustrating its practicability and effectiveness. {The performance is comparable to the ER doctors' clinical impressions (CI). This is also of great significance considering the doctors generally will refer to the patients' CT results, vital signals, and EHR for past stroke history. From Fig.~\ref{fig:roc_base}, the proposed \textit{DeepStroke} achieves close (and slightly better) performance. For better comparison, we report results when the specificity and sensitivity of \textit{DeepStroke} aligned with ER CI in Table~\ref{tab:ci}}.

\begin{table}[htbp]
\small
\centering
\renewcommand\arraystretch{1.2}
\caption{{Model and baseline performance compared with ER doctor's Clinical Impression (ER CI). }}
\resizebox{\linewidth}{!}{
\begin{tabular}{rccc}
\toprule
\textbf{Baselines} & \textbf{Accuracy (\%)} & \textbf{Specificity (\%)} & \textbf{Sensitivity (\%)}  \\\midrule
Clinical Impression (CI)                              & {69.23}   & {{51.61}}        & {76.10}         \\  
\textbf{\textit{DeepStroke}} (Spec. Aligned)    & {\textbf{71.04}}   & {{51.61}}        & {\textbf{78.61}}      \\  
\textbf{\textit{DeepStroke}} (Sens. Aligned)    & {69.68}   & {\textbf{53.22}}        & {76.10}      \\  
\bottomrule
\end{tabular}}\label{tab:ci}
\end{table}

{Summarizing Table~\ref{tab:baseline}, Table~\ref{tab:ci}, and Fig.~\ref{fig:roc_base}, we conclude that the proposed \textit{DeepStroke} is powerful in performing pre-hospital stroke screening and is able to outperform traditional stroke triage. It achieves comparable or better performance with much less available information.}


With new patients continually being added to the cohort, our dataset is becoming more and more diverse and even more challenging for the clinicians (for the original dataset, clinicians had 72.94\% accuracy, 77.78\% specificity, and 70.68\% sensitivity). We infer that this is due to the addition of a number of hard cases, where the patterns for stroke are too subtle for the clinicians to capture. Even so, when we align specificity, the proposed method still outperforms the clinicians. ER doctors tend to rely more on the speech abilities of the patients and may have difficulty in cases with too subtle facial motion incoordination. We infer that the video module in our framework can detect those subtle facial motions that doctors can neglect and complement the diagnosis based on speech/audio. The drop in specificity is regarded as permissible compared to the improvements in sensitivity because, in stroke screening, failing to spot a patient with stroke (false-negative) will result in very serious results. 
\subsubsection{Holdout Experiment}
{Aside from the cross-validation experiments, to ensure the clinical significance of our proposed methods, we conducted a holdout experiment. Cases in our dataset are ordered by their time of data acquisition. To maximally resemble the clinical situation, among the 221 collected cases, we set the first 168 cases for training/validation and the subsequent 53 cases for testing. This keeps the ratio of the testing set to be 25\% of the dataset, while we can ensure the testing set has a similar class ratio to the training/validation set. In the training/validation set, we have 121 stroke cases and 47 non-stroke cases resulting in a ratio of 2.57; in the testing set, there are 38 stroke cases and 15 non-stroke cases, giving a ratio of 2.53. In this holdout study, we test the best models acquired from 5-fold cross-validation (with training/validation set) on the \textit{unseen} hold-out testing data. We refer to AUC as the saved model performance benchmark on the testing data. Because each of the five folds will save the best model, we report the mean, standard deviation, and range of the AUC across these models. We also plot the cross-fold AUC for our proposed methods and baselines to demonstrate the stability of our method. Note that due to the apparent low performance of the state-of-the-art video baselines, we omit I3D and SlowFast in the holdout experiment.}
\begin{table}[htbp]
\footnotesize
\centering
\renewcommand\arraystretch{1.1}
\setlength{\tabcolsep}{11pt}
\caption{{Results of the hold-out experiment. The metrics are calculated based on the testing AUC of the five saved models from each fold on the aforementioned testing set. SD: standard deviation. $\uparrow$ indicate higher the value, better the performance, and $\downarrow$ means the opposite. The best values are in bold.}}
\begin{tabular}{rlll}
\toprule
 & \multicolumn{1}{c}{\textbf{Mean$\uparrow$}} & \multicolumn{1}{c}{\textbf{SD$\downarrow$}} & \multicolumn{1}{c}{\textbf{Range$\uparrow$}} \\\midrule
Module $\Gamma_\text{a}$ (Audio) & 0.4631                   & 0.1389                 & {[}0.2263, 0.6386{]}        \\
Module $\Gamma_\text{v}$ (Video) & 0.5242                   & \textbf{0.0320}                 & {[}0.4895, 0.5649{]}        \\
MMDL                             & 0.6417                   & 0.0475                 & {[}0.6018, 0.7333{]}        \\
\textit{\textbf{DeepStroke}}     & \textbf{0.7214}             & 0.0368                 & \textbf{{[}0.6772, 0.7895{]}}       \\
\bottomrule
\end{tabular}\label{tab:holdout}
\end{table}

\begin{figure}[ht!]
    \centering
    \includegraphics[width=.9\linewidth,trim={0.9cm 0.2cm 1.2cm 0.5cm}]{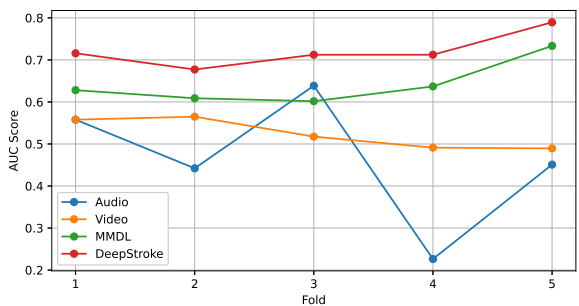}
    \caption{{Cross-fold hold-out testing AUC plot for the proposed \textit{DeepStroke} and other baselines. }}
    \label{fig:holdout}
\end{figure}

{As shown in Table.~\ref{tab:holdout}, the \textit{DeepStroke} framework can achieve stable and high performance in our stroke screening task. A steady improvement from our previous work (MMDL) is observed. To better present the results, we plot the testing AUC of different models in Fig.~\ref{fig:holdout}. }

{The testing performance from any of the folds generated with the \textit{DeepStroke} is stable and high. With audio alone, the test performance has seen large variations across different fold models. This is not presented in the cross-validation experiment above. In some folds (\textit{i.e.,} fold \#4 in Fig.~\ref{fig:holdout}), the saved model was performing the best in the validation set but is giving a very low AUC score on the testing set. We infer that merely relying on the patient's speech audio is insufficient to recognize stroke patterns, as the audio input includes a large number of unused patterns and noises. The video data, on the other hand, performs relatively stable yet has been unable to recognize non-stroke cases. The collaboration of the audio module and the video module, as both the previous work and the proposed \textit{DeepStroke} demonstrates, achieves noticeable performance gain. Furthermore, with the proposed improvements over the previous work, \textit{DeepStroke} can perform stroke screening with higher stability and better performance.}
\subsection{Ablation Study}\label{sec:abl}
\subsubsection{Model designs}\label{S5.3.1}
To evaluate the effectiveness of different designs in our proposed model,
we perform the following ablation studies. {Note that the reported statistics are based on the 5-fold cross-validation results on the same dataset stated above. Here, we denote \textit{DeepStroke} as the full model that introduced lateral connection, adversarial traning, and transfer learning to the network introduced in our previous work.}
\begin{itemize}
    \item {\textbf{Model without transfer learning (w/o transfer):} We demonstrate the value of our transfer learning setup by contrasting it with a baseline model that takes the same network design as the \textit{DeepStroke} but does not freeze parameters. The baseline model discards 1) FairFace as face video network backbone pre-training and 2) ImageNet as audio network pre-training, and was trained from scratch.}
    \item \textbf{Model without lateral fusion {(w/o lateral)}:} The lateral connection we designed for the two branches was updated from the late fusion scheme in our prior work to connect the two branches at different levels. An experiment was made to only use the late fusion as a comparison to the full proposed model to show the power of the lateral fusion.
    \item \textbf{Model without adversarial training {(w/o adversarial)}:} The adversarial training scheme aims to extract stroke-related identity-free features. We train the same model without the adversarial scheme to see the improvements.
    \item \textbf{Frame difference {(Conv-diff)}:} In the proposed method, we set the two consecutive frames to form a difference at the very beginning of the network, but such subtraction can be made elsewhere--in the middle of the network or at the back of the network (\textit{i.e.}, the deep features generated from the framework). An extra experiment is conducted on the frame subtraction scheme that takes the difference of features after one Conv layer as used in our prior work.
    \item \textbf{Discriminator feature input stage {(Feature \#)}:} When adversarial training for distilling the subjects' identities, our network takes the encoded features after all four blocks of ResNet. We have other stages where features are extracted for the adversarial training, (\textit{i.e.}, after the first block, after the third block, and after the third layer). 
    \end{itemize}
\begin{table}[htbp]
\centering
\renewcommand\arraystretch{1.2}
\setlength{\tabcolsep}{5pt}
\caption{Ablation study results. The first block of rows ablates the lateral connection we designed over the model components, the second block ablates the other choice of making frame pair difference, and the third block ablates other choices of extracting encoded features for the discriminator. {Experiments share the same setup as the cross-validation experiments in Sec.~\ref{sec:cv5} and parameters are tuned separately. }}
\resizebox{\linewidth}{!}{
\begin{tabular}{rcccc}
\toprule
\textbf{Baselines} & \textbf{Accuracy (\%)} & \textbf{Specificity (\%)} & \textbf{Sensitivity (\%)} &  \textbf{AUC} \\\midrule
w/o transfer & 62.44& 37.10& 72.33& 0.5611\\
w/o lateral &     71.49           &      37.10    &   84.91    &   0.6881       \\
w/o adversarial &      68.87             &    17.78         &  90.57   &   0.5568      \\\midrule
Conv-Diff &  69.78   &  25.80  &  85.53  &  0.6680   \\\midrule
Feature \#1 &  69.23  & 30.64 & 84.28 &  0.6679 \\
Feature \#2 & 69.23  &25.80  & 86.16   & 0.6690   \\
Feature \#3 &  69.23   & 16.13   &  89.94   &   0.6678    \\\midrule
\textbf{\textit{DeepStroke}} &  72.40 &  32.26  & 88.05 & \textbf{0.7163}     \\
\bottomrule
\end{tabular}}\label{tab:ablate}
\end{table}

From Table~\ref{tab:ablate}, we could conclude that the lateral connection, transfer learning, and adversarial training have been improving the performance of the framework as expected. The lateral connection improves the overall statistics with 0.0282 in AUC, indicating a better learning outcome that may result from better balanced two-branch learning dynamics. By applying transfer learning, we noticed a clear performance gain in terms of sensitivity of 15.72\% and a 0.1552 improvement in AUC. The FairFace pre-training reduces the potential biases from facial attributes so that the network has less reference to such features when making a decision. The adversarial training also drastically improved the specificity (14.48\% as shown) and resulted in a better learning result (0.1483 improvements in AUC), and we believe the features for stroke are more clearly revealed with this training scheme. Our choice of features out of different network layers also demonstrated the effectiveness of using final concatenated features for the use of a discriminator.

\subsubsection{Model fairness}

We also examined the discrepancy between different genders, races/ethnicity, and age groups in our dataset. Specifically, we examine the following attribute groups in Table~\ref{tab:attr} and report the validation results within the 5-fold cross-validation experiment result on the proposed model:
\begin{itemize}
    \item \textbf{Gender:} Previous studies point out that gender discrepancy is noticeable for stroke: women show unique patterns that are often neglected~\citep{colsch2018unique}. The probability of men getting misdiagnosed with stroke is lower than women~\citep{newman2014missed}. To compare how clinicians and our model perform regarding this discrepancy, we report results on different gender groups.
    \item \textbf{Age:} The risk for stroke increases with age. According to national statistics, the majority of patients hospitalized for stroke are over 65 years old~\citep{hall2012hospitalization}. This prior is not clearly shown in our dataset, partly because some older patients are having more serious conditions and are not enrolled in this study. 
    \item \textbf{Ethnicity:} According to~\cite{trimble2008stroke}, the incidence of stroke in Hispanics is significantly higher than in Non-Hispanic whites. Hispanics are also believed to have a higher rate of stroke recurrence~\citep{sheinart1998stroke}. Note that due to the relatively limited Hispanic participants in our current version of the dataset, the comparison may not be indicative enough for the discrepancy.
    \item \textbf{Race:} African Americans have nearly twice as high a risk of first stroke compared to others~\citep{virani2020heart} and the highest mortality rate~\citep{cdc2}. African Americans also receive less evidence-based care and have a higher risk of stroke recurrence~\citep{schwamm2010race}. Another potential issue is concerned with the biases in current face-related computational models that are mainly trained on image/video samples from non-African-American participants. 
\end{itemize}

Table~\ref{tab:attr} shows that the validations on most of the attribution groups result in a relatively stable AUC (variations are expected considering the relatively small number of samples). For the attribute of gender and ethnicity, the variation among different groups can be regarded as reasonable. The results on African American subjects raise some concern as the AUC falls to 0.5580, indicating that the model trained on the complete dataset is not picking up the African American cases very well. This is likely due to the larger imbalance of positive/negative cases among African American groups compared to the complete dataset (54 stroke cases, 18 non-stroke cases). A further ablation study is conducted to train the framework with different setups on other cases and validate on African American cases and the result is shown in Table~\ref{tab:aa}.
\begin{table}[htbp]
\small
\centering
\renewcommand\arraystretch{1.2}
\setlength{\tabcolsep}{4.5pt}
\caption{Validation results for African American cases when training the framework on other cases. {The experiments are trained on 149 other cases while 72 African American cases are used as validation data. Best test performance on African American test set in 20 epochs is reported.}}
\resizebox{\linewidth}{!}{
\begin{tabular}{rcccc}
\toprule
\textbf{Experiment} & \textbf{Accuracy (\%)} & \textbf{Specificity (\%)}& \textbf{Sensitivity (\%)}&  \textbf{AUC} \\\midrule
\footnotesize{\textit{DeepStroke}} &70.83& 50.00 & 77.77 & 0.6967 \\
\footnotesize{w/o FairFace} &66.66& 38.89 & 75.93 & 0.6821 \\
\footnotesize{w/o Transfer Learning} & 63.88 & 61.11  & 64.81  & 0.6255  \\
\bottomrule
\end{tabular}}\label{tab:aa}
\end{table}

{In this hold-out experiment, we notice that the model performance is respectable when trained with no African American cases but validated only on African American cases. Using the full \textit{DeepStroke}, the validation performance is at 0.6967 AUC. We then replaced the FairFace pre-training with ImageNet pre-training and obtained a slightly worse result with 0.6821 AUC. If trained from scratch, \textit{i.e.,} without transfer learning, the model performance dropped vastly to 0.6255 AUC. From the experimental results in Table~\ref{tab:aa}, we can conclude that, by adopting the transfer learning scheme and FairFace pre-training, \textit{DeepStroke} can better reduce facial attribute-related biases and improve the generalizability of the method. }

\begin{table*}[ht!]
\small
\centering
\renewcommand\arraystretch{1.17}
\setlength{\tabcolsep}{11pt}
\caption{Validation results for different attribute groups. The number in the bracket indicates the total number of cases in the group. Some subjects opt-out for ethnicity or race questions and are excluded. Model: \textit{DeepStroke}, CI: Clinical Impression. Values in \textcolor{red}{red} are considered to yield fairness concerns. }
\begin{tabular}{rrcccccccc}
\toprule

\multirow{2}{*}{} &\multirow{2}{*}{} & \multicolumn{2}{c}{\textbf{Accuracy (\%)}} & \multicolumn{2}{c}{\textbf{Specificity (\%)}} & \multicolumn{2}{c}{\textbf{Sensitivity (\%)}} & \multicolumn{2}{c}{\textbf{AUC}} \\ \cline{3-10} 
                                       \textbf{Group}                  & \multicolumn{1}{c}{\textbf{Attribute}}   & Model         & CI            & Model           & CI             & Model           & CI             & Model       & CI         \\  \midrule
\multirow{2}{*}{Gender} & Male (108)                                                    & 75.00         & 66.39         & 34.38           & 60.00          & 92.10           & 69.05          & 0.6772       & ---      \\
 & Female (113)                                                  & 70.80         & 66.11         & 25.00           & 48.49          & 83.15           & 72.72         & 0.6905       & ---      \\ \midrule
\multirow{2}{*}{Age} & $\leq$ 65 y/o (120)                                           & 72.50         & 67.67         & 40.00           & 54.76          & 83.33           & 73.63          & 0.7048       & ---      \\ 
 & $\ge$ 65 y/o (101)                                            & 73.27         & 64.49         & \textcolor{black}{19.23}           & 53.84          & 92.00           & 67.90          & 0.6728       & ---     \\ \midrule
\multirow{2}{*}{Ethnicity} & Hispanic (16)                                                & 75.00         & 88.23         & 40.00           & 75.00          & 90.90          & 92.31          & 0.8545       & ---      \\ 
 & Non-Hispanic (198)                                           & 72.22         & 65.27         & 28.57           & 55.17          & 86.58           & 68.99          & 0.66.15       & ----      \\ \midrule
\multirow{2}{*}{Race} & African American (72)        & 69.44         & 61.84         & 23.08           & 45.00          & 79.66           & 67.86          & \textcolor{red}{0.5580}       & ---      \\ 
 & Other (149)                                                  & 74.50         & 68.29         & 32.56           & 58.33          & 91.51           & 72.41          & 0.7411       & ---      \\ \midrule
\multicolumn{2}{c}{All subjects (221) }                                          &    72.40      &    69.23      &    32.26        &    51.61       &     88.50       &     76.10      &    0.7163    & ---      \\ 
\bottomrule
\end{tabular}\label{tab:attr}
\end{table*}

Another potentially biased attribute group is age. Age is a known factor for the risk of stroke and with higher age, the probability of stroke is greatly increased. Both the model and clinical impression have a rather large discrepancy in specificity between the two age groups, and cases younger than 65 years old are much higher in specificity. To further look into and attempt to understand this discrepancy, we first train on younger cases and validate on older cases, and then do the reverse. The results are shown in Table~\ref{tab:age}.
\begin{table}[htbp]
\small
\centering
\renewcommand\arraystretch{1.2}
\setlength{\tabcolsep}{4.5pt}
\caption{Ablation study on age attribute. The first row is trained on younger cases ($\leq$ 65 y/o) and tested on older cases ($\geq$ 65 y/o), and the second row is trained on older cases and trained on younger cases.}
\resizebox{\linewidth}{!}{
\begin{tabular}{rcccc}
\toprule
\textbf{Experiment} & \textbf{Accuracy (\%)} & \textbf{Specificity (\%)}& \textbf{Sensitivity (\%)}&  \textbf{AUC} \\\midrule
$\leq$ 65 y/o &68.31& 30.77 & 81.33 & 0.6203 \\
$\geq$ 65 y/o &68.33& 38.89 & 80.95 & 0.6257 \\
\bottomrule
\end{tabular}}\label{tab:age}
\end{table}

The result shows a similar classification accuracy at 0.62 AUC and is lower than training on all data, which partly indicates that the network may be able to extract useful age information from the patients and use it as a feature for prediction. We infer that the relatively lower specificity and high sensitivity among the age group over 65 years old is also a consequence of heavier class imbalance (79 stroke cases, 26 non-stroke cases), and the network prefers to give a positive prediction on patients in the higher age band. This could reduce the risk of missing a real stroke case in the ER and could potentially save the patient. 

\section{Discussion}
We analyze the running time of the proposed approach. The recording runs for a minute, the extraction of audio and generation of spectrograms takes an extra minute, and the video processing is completed in three minutes. The prediction with the deep models can be achieved within half a minute on a desktop with a mid-level GPU (NVIDIA GTX1070). Therefore, the evaluation process takes no more than six minutes per case. More importantly, the process is almost running at zero external cost and would be contactless, not harming the patients with equipment or by radiation. Considering a complete MRI scan will take more than an hour to perform, a specialized device to run, and hundreds of dollars charged, the proposed method is ideal for performing both cost and time-efficient pre-hospital stroke assessments in an emergency setting. 

{It is worth mentioning that although our proposed method is demonstrated to outperform triage stroke screening performance by a margin and is comparable to or better than ER doctors who have access to richer patient data, its use should be limited to that of an AI assistant during the ER triage stage to help to detect non-obvious strokes and mitigate the risk of misdiagnosing. The CT scan should be taken as the next step to identify the stroke sub-types and estimate damaged regions for the reference of doctors in performing interventions. }

We expect that our proposed approach will be clinically relevant and can be deployed effectively on smartphones for fast and accurate assessment of stroke by ER doctors, tele-stroke neurologists, at-risk patients, or caregivers. If the approach is further optimized and deployed onto a smartphone, we can perform the spatiotemporal face frame proposal and speech audio processing on the phone. Cloud computing can be leveraged to perform the prediction in no more than a minute after the frames are compressed and uploaded. In such a case, the total time for one assessment should be within a few minutes. Moreover, with minimal user education required, such a framework can allow for the patients' self-assessments even before the ambulance arrives. With different labeling of data, the pipeline is also valuable in the screening of other oral-facial neurological conditions. 

We also noticed that when clinicians are performing stroke screening in the ER, they would consider race, ethnicity, and age as factors, and they may also refer to the patients' EHR for previous stroke records. Such information is missing from our dataset. While we would like to find patterns directly from facial motions and speaking voices, these factors are certainly beneficial to be adopted as auxiliary information that may guide the framework to project different probabilities of stroke for different demographic groups referring to the prior distributions. 

Through our experiment, we noticed that the dataset we have been collecting for the pilot study purposes is still preliminary and small, especially for the evaluation of various demographic subgroups. Also, with a limited number of cases available, the dataset could be insufficient to cover all possible patterns of stroke in facial motions and speech. The limitation also leads to the performance discrepancy in the African American cases and among age-related attribute groups. We are working towards a large-scale clinical trial that targets about 1,000 cases to evaluate the proposed method and further improve its robustness. When enough data is available, we would also pick a verified set of held-out test data to cover expected variations and use it as the public evaluation standard for this task.

{To compensate for the relatively small data size, while we have been trying to expand our dataset, we also considered two data augmentation schemes and attempted to apply them to the framework, including:
\begin{itemize}
    \item We recruited Amazon MTurk independent contractors to perform the same speech tasks and use these data to augment the non-stroke cases. For future study purposes, we collected more than enough data to balance the minority class. We trained an audio ResNet-18 (audio module) based on spectrograms and noticed higher testing performance and better stability of the augmented audio module in the hold-out experiment. However, when the balanced-data-trained models were set as transfer learning starting points for the audio sub-network, testing AUC of the whole model (including both audio- and video- modules) drastically dropped. We suspect this is because audio pre-trained model augmented by normal audio makes the whole model less generalizable.
    \item We used one of the best-performing facial motion transfer algorithms~\citep{wang2021facevid2vid} and attempted to transfer facial motions from non-stroke patients to public online faces, and pair the transferred facial movies with the collected MTurk audio to generate synthetic negative cases. However, the addition of such cases did not help improve the network performance. We noticed that the motion transfer constantly fails to preserve stroke-indicative motions and introduces more artifacts, mainly due to wrinkles and other uncommon facial patterns. 
\end{itemize}
We will further investigate other possibilities of data augmentation for improving the framework's performance}

We recognize other limitations of our work. First, the spatiotemporal facial frame proposal method is preliminary and is unable to eliminate all non-facial motions. This induces errors in the captured facial motion features. Also, the state-of-the-art 2D facial landmark tracking and pose estimation algorithms still induce considerable error in the pipeline. A plan is to collect 3D data instead of 2D so that we could achieve accurate alignment and reconstruction of subjects' faces. With an accurate mesh representation of the subjects' facial motions, we could model the motions of different facial regions and correlate them to different areas of brain lesions. The representation could also help reduce color bias and naturally remove identity from subjects, which is promising. Second, the computational workload of the whole pipeline is still heavy for mobile deployment. In conditions with low network bandwidth, the transfer of collected data could be a serious bottleneck and hinder the efficiency of the screening process. Future work would need to address these issues and improve the framework for even better clinical performance.

{As a future work, we intend to develop \textit{DeepStroke+} that resolves the existing issues of our current framework and functions as a full stroke diagnosis AI assistant to span the stroke triage till discharge and treatment. \textit{DeepStroke+} will be based on the much larger dataset collected through the clinical trial and adopt cloud/mobile computing for efficient, mobile assessment. The anticipated model will be taking 3D facial data uploaded with mobile portals and linked to the patient's EHR, demographics, vitals, and CT report as multi-modal input. \textit{DeepStroke+} will be developed as a multi-task network to support the additional decision of stroke subtypes and lesion regions and provide detailed reports to the ER clinicians. }

\section{Conclusion}
In this work, we presented a novel multi-modal deep learning framework, \textit{DeepStroke}, for on-site clinical triage of stroke in an ER setting. Our framework can perform accurate and efficient stroke screening based on the abnormalities in the patient's speech ability and facial muscular movements. We constructed a dual branch deep neural network for the classification of patient facial video frame sequences and speech audio as spectrograms to capture subtle stroke patterns from both modalities. Experiments on our collected clinical dataset with real, diverse, ``in-the-wild'' ER patients demonstrated that the proposed approach not only outperforms the traditional stroke triage with a 10.94\% higher sensitivity rate and 7.37\% higher accuracy when specificity is aligned but also outperforms well-trained ER clinicians with more information available. Also, ablation studies validated the value of our multimodal lateral fusion method, transfer learning pipeline, and adversarial training scheme, which collaboratively improve our proposed model from our prior work. The \textit{DeepStroke} has also been verified to be efficient and provide a screening result for the reference of clinicians in minutes. 

Before a clinical deployment, we will continue to conduct the clinical trial, enlarge the dataset, improve the efficiency of the pipeline, and develop the current method into a 3D mobile framework. Future efforts may also try to address the potential biases in the current model and apply the method to other neurological conditions. 

\section*{Acknowledgments}
T. Cai, H. Ni, M. Yu, X. Huang, and J.Z. Wang were supported by the College of Information Sciences and Technology at The Pennsylvania State University. T. Cai, K. Wong, and S.T.C. Wong were supported by the T.T. and W.F. Chao Foundation, the John S. Dunn Research Foundation, and The Scullock Foundation. The computation was supported by the NVIDIA Corporations GPU Grant Program and the Extreme Science and Engineering Discovery Environment (XSEDE), which is supported by the National Science Foundation grant number ACI-1548562. The authors would like to thank Amber Criswell for her help in patient enrollment and Xiaohui Yu for mobile app development. The authors would also like to acknowledge the comments and constructive suggestions from the reviewers and the associate editor.

\bibliographystyle{model2-names.bst}\biboptions{authoryear}
\bibliography{ref}
\end{document}